# Large Language Models for Accurate Medical Chart Abstraction: Enabling Scalable and Secure Artificial Intelligence Deployment in Stroke

Zhusi Zhong[1,2], Carl M. Porto[1], David Hong[1], Miranda Gonzales[1], Revyn Kim[1], Gnaneswari Karayi[1], Lulu Bi[1], Joshua R. Feler[1,4], Elias Shaaya[1], Scott Collins[1], Liqi Shu[1], Grayson Baird[1,2], Mahesh Jayaraman[1,2], Shadi Yaghi[1], Zhicheng Jiao[1,2], Dylan N. Wolman[1,3]

[1] Warren Alpert Medical School of Brown University, Providence 02903, USA

[2] Department of Radiology, Brown University Health, Providence 02903, USA

[3] Department of Interventional Radiology, Brown University Health, Providence 02903, USA

[4] Department of Neurosurgery, Brown University Health, Providence 02903, USA

Correspondence: Dylan N. Wolman; dylan_wolman@brown.edu

### Summary statement

Prompted LLMs enable accurate and scalable extraction of structured clinical variables from neurointerventional stroke procedure reports with key radiology findings, supporting automated chart abstraction for large-scale research and registry curation.

### Key Points

We developed and systematically evaluated an automated chart abstraction framework leveraging large language models (LLMs) for extraction of eight clinically relevant variables from 2,416 free-text neurointerventional procedure reports across three hospitals.

Twenty-two state-of-the-art, instruction-tuned LLMs spanning diverse architectures, parameter scales, and biomedical adaptations, were systematically benchmarked to assess their accuracy and scalability in structured clinical variable extraction.

Expert adjudication of cases with discrepant predictions demonstrated that LLaMA3.3-70B model, with chain-of-thought prompting, outperformed non-expert staff (without formal clinical training) on variables requiring contextual inference in 7 of 8 categories, and achieved accuracy comparable to that of junior medical students.

## Abstract

BACKGROUND AND PURPOSE: Medical chart abstraction plays a critical role in clinical research and quality monitoring by transforming unstructured narratives in the procedure reports into structured variables for large-scale analysis. To develop and evaluate a prompting-based large language model (LLM) framework for automated extraction of structured clinical variables from neurointerventional procedure reports in patients with acute ischemic stroke (AIS) due to large-vessel occlusions (LVO).

MATERIALS AND METHODS: This retrospective study included 2,416 free-text neurointerventional acute stroke intervention (thrombectomy) reports with key radiology findings from three hospitals. Eight clinically relevant variables were annotated by hospital staff (without formal clinical training) and used as the non-expert reference standard. Twenty-two instruction-tuned open-source LLMs (LLaMA, Qwen, Gemma, etc) were evaluated across architectures, sizes, and biomedical adaptations using two prompting strategies: Quick Response and Chain-of-Thought (CoT). Model performance was benchmarked against non-expert staff annotations and medical expert ratings. Extraction accuracy, latency, and agreement with expert adjudication were assessed.

RESULTS: LLaMA3.3-70B achieved the highest overall accuracy (94.8%). CoT prompting improved performance on inferential variables (e.g., site of occlusion), while Quick Response was optimal for directly stated procedural fields (e.g., stent placement). Expert adjudication confirmed that LLaMA3.3-70B outperformed non-expert annotations in 7 of 8 variables and matched junior medical students. Annotation accuracy increased with clinical experience, and AI predictions were more closely aligned with expert interpretations than with those of non-expert staff, especially for structured variables like IV tPA, TICI Post, and NIHSS.

CONCLUSIONS: Prompted LLMs can accurately and scalably extract critical clinical information from neurovascular radiology reports without custom preprocessing, supporting integration into retrospective research pipelines and automated stroke registry curation.

### Abbreviations

CoT = Chain-of-Thought; CTA = computed tomography angiography; LVO = large vessel occlusion; ICA = internal carotid artery; IV tPA = intravenous tissue plasminogen activator; LLM = large language model; NLP = natural language processing, NIHSS = National Institutes of Health Stroke Scale; NASCET = North American Symptomatic Carotid Endarterectomy Trial; TICI = Thrombolysis in Cerebral Infarction (score).

### Keywords

Medical Chart Abstraction, Stroke Triage, Large Language Models, CTA Head and Neck, Multi-Hospital AI Deployment, Structured Data Extraction, Radiology NLP, ELVO, Thrombectomy Support, Multi-Modality Imaging

## Study context

Previous Literature: Prior studies have explored natural language processing approaches for extracting information from radiology reports, including rule-based systems, traditional machine learning, and more recently large language models. However, most investigations have focused on impression-level summarization or single-task extraction, often using small cohorts or cloud-based proprietary models. Evaluations of structured variable extraction from neurointerventional stroke procedure reports remain limited, particularly in multi-institutional settings with local deployment constraints and direct comparison to human annotators of varying clinical expertise.

Key Findings: Across 2,416 neurointerventional stroke reports, large instruction-tuned LLMs—particularly LLaMA3.3-70B with chain-of-thought prompting—accurately extracted structured clinical variables, outperforming non-expert staff on inference-dependent fields and achieving performance comparable to junior medical trainees.

Knowledge Advancement: This study establishes a clinically grounded benchmark for LLM-based chart abstraction in stroke care, demonstrating that locally deployed LLMs can reliably extract diagnostic and procedural variables from real-world reports, while highlighting the roles of prompting strategy, model scale, and expert oversight in scalable clinical deployment.

## Introduction

Accurate extraction and aggregation of diagnostic and procedural information from radiology reports at scale may be beneficial for optimizing stroke systems of care, particularly for patients with acute ischemic stroke due to large-vessel occlusions (LVO)[1-3]. In addition, automated extraction of clinical and procedural data may help streamline labor-intensive processes such as data collection for stroke center accreditation (e.g., Comprehensive Stroke Center [CSC] certification by The Joint Commission) and ongoing quality improvement or outcomes research [4]. As illustrated in Figure 1, the structured data derived from free-text reports support a wide range of downstream tasks, including retrospective cohort curation, quality assurance benchmarking, and system-wide performance monitoring[5]. However, current chart abstraction pipelines rely on manual review by trained personnel, a process that is time-consuming, labor-intensive, and prone to inter-rater variability[6, 7].

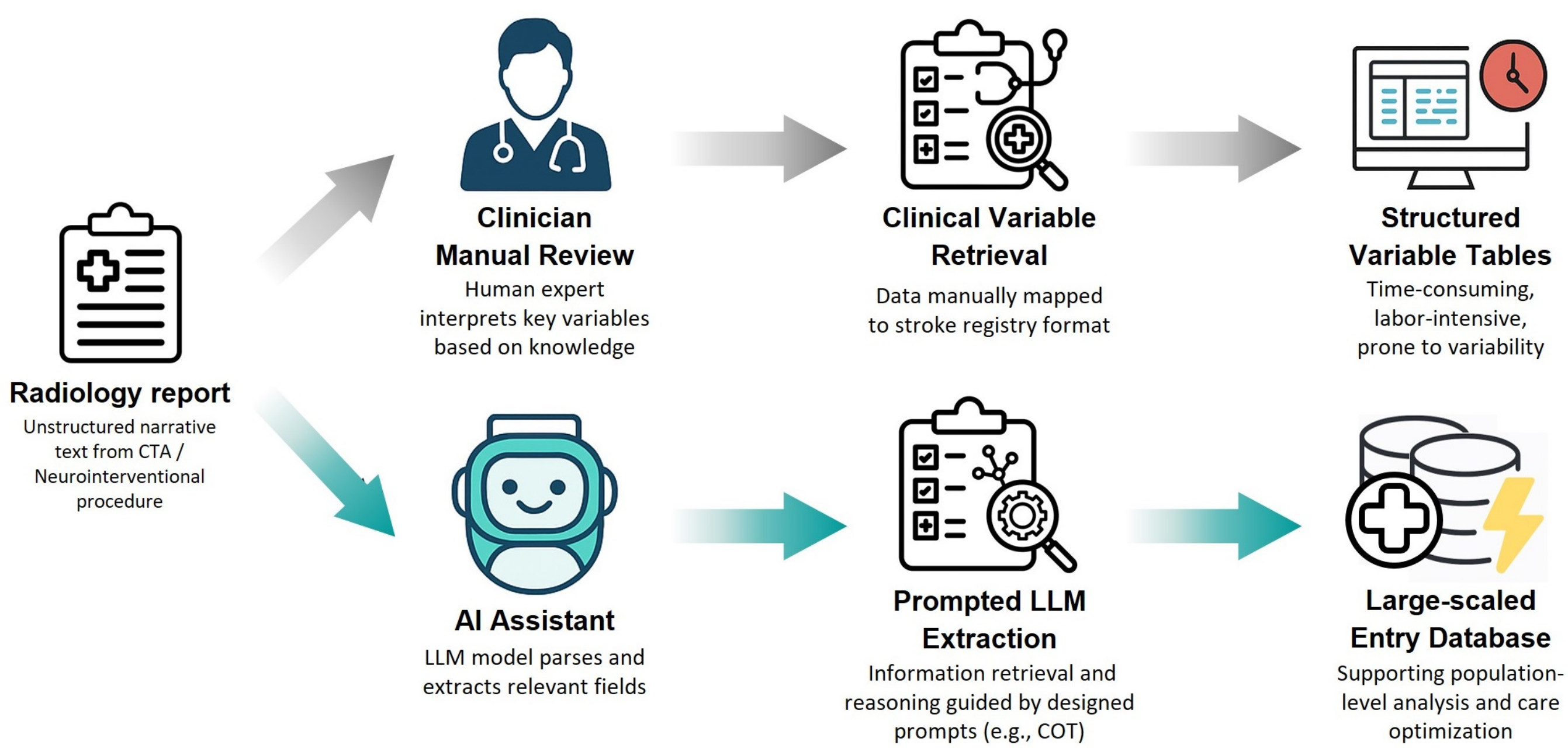

Figure 1. Overview of traditional vs. LLM-assisted clinical variable abstraction pipelines. In the conventional workflow (top), clinicians manually document reports, which are later reviewed by staff for variable abstraction—a time-consuming and labor-intensive process. The LLM-assisted pipeline (bottom) enables direct extraction of structured variables from radiology narratives, streamlining data abstraction for real-time analytics and large-scale registry integration.

Recent advances in large language models (LLMs), offer new opportunities to automate this process[8-11]. State-of-the-art LLMs have achieved strong performance in biomedical tasks such as named entity recognition and relation extraction[11, 12]. When integrated into radiology informatics workflows, LLMs may facilitate real-time, scalable clinical data abstraction without the need for custom rules or templates[13].

Despite their potential, the application of LLMs to structured variable extraction from neurointerventional procedure reports remains underexplored. Related studies have been constrained by small cohorts and online models, limiting clinical scalability and privacy compliance [14]. More prior works have focused primarily on impression-level summarization or general-purpose clinical text parsing, with limited evaluation in the context of acute stroke reporting[15, 16].

In this study, we systematically evaluate 22 instruction-tuned LLMs for extracting 8 clinically relevant variables from 2,416 acute stroke intervention procedure reports. We examine two prompting strategies (Quick Response vs. Chain-of-Thought), and assess model performance across sizes, architectures, and biomedical adaptations. To contextualize LLM accuracy, we benchmark AI predictions against human annotations from both clinical experts and trainees, providing a real-world assessment of feasibility for LLM-based abstraction in stroke care.

## Materials and Methods

### Cohort and Variable Collection

A total of 3,624 consecutive acute stroke cases were identified from three Brown University Health-affiliated hospitals between July 17, 2015, and December 11, 2024. Eligible cases were adults (≥18 years) who underwent head and neck CTA followed by endovascular thrombectomy for suspected or confirmed LVO. After excluding cases without a valid medical record number (MRN) or with missing reports, 2,416 neurointerventional procedure reports remained and were used for validation of LLM-based medical chart abstraction (as cohort flow diagram shown in Figure 2). The study was approved by the Institutional Review Board (IRB) of Brown University Health (IRB protocol number 2226291-11) with a waiver of informed consent due to its retrospective design and use of a prospectively maintained registry. All reports were de-identified, with no external transfer of protected health information, in full compliance with HIPAA regulations and institutional policies.

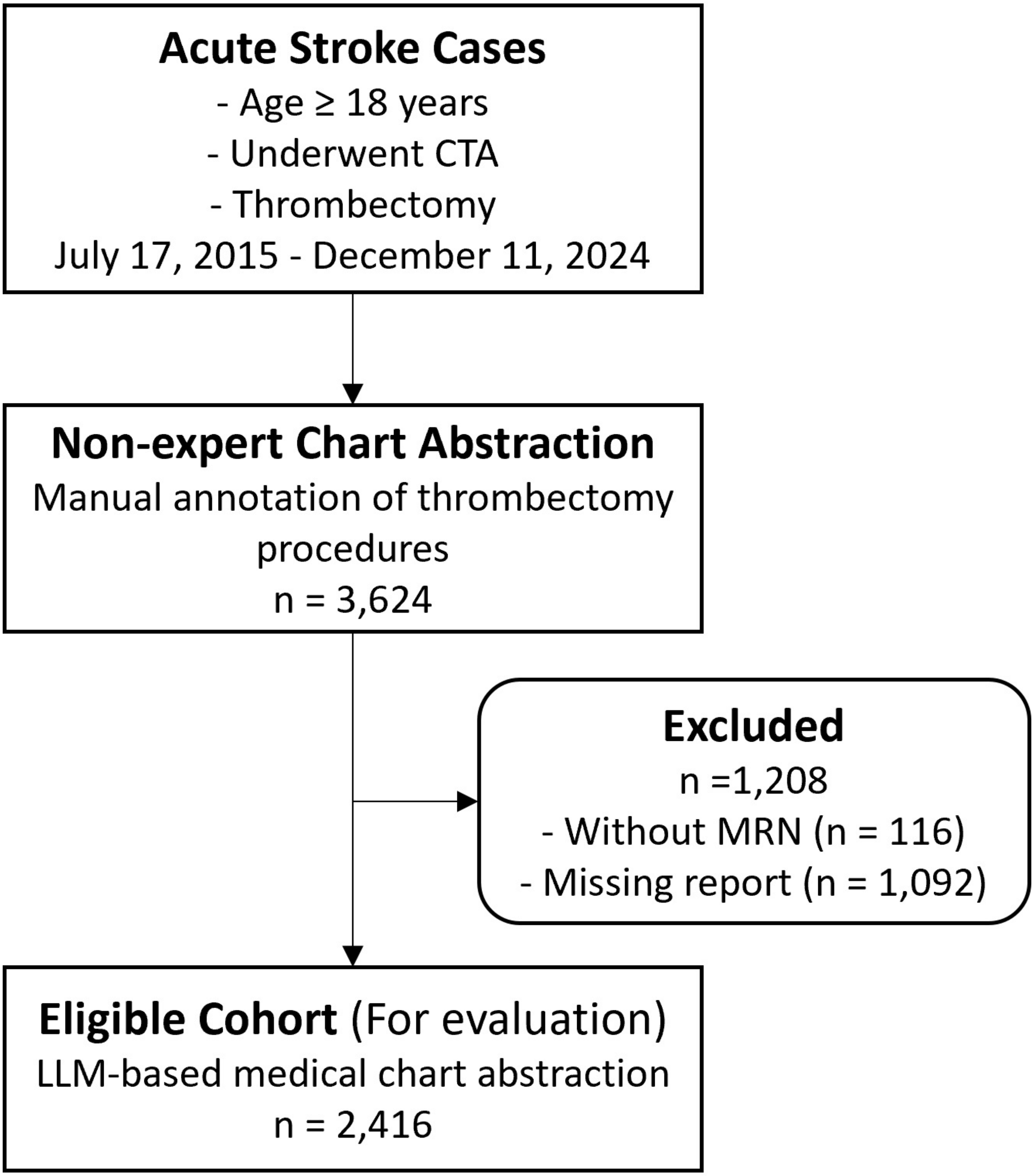

Figure 2. Identification and selection of neurointerventional stroke procedure reports. Flow diagram depicting cohort identification, non-expert chart abstraction, exclusion of incomplete cases, and final report selection used for validation of LLM–based clinical variable extraction

Within the institutional health network, the neurointerventional report adhered to a standardized documentation protocol and included detailed procedural elements such as access route (e.g., radial or femoral), catheters/devices used (e.g., balloon guide catheters, stent retrievers), number of thrombectomy passes, sedation methods, administered medications (e.g., heparin, verapamil), contrast volume, and radiation dose (an anonymized report sample is provided in Table S.2 of Supplemental Data ). CTA imaging findings were described in detail, including occlusion site, collateral status, and reperfusion outcomes. Post-procedural angiographic results were graded using the modified Thrombolysis in Cerebral Infarction (mTICI) scores, and vessel stenosis was quantified using NASCET criteria. Reports concluded with a physician-signed impression and disposition.

Building on the standardized structure of the neurointerventional reports described above, eight key clinical variables were systematically selected and extracted to support structured medical chart abstraction and downstream clinical analysis. Within the procedure reports, site of occlusion and NIHSS are primarily derived from the CTA radiology findings and clinical assessment sections, whereas IV thrombolysis status, post-procedure TICI grade, carotid stenting, carotid angioplasty, intracranial stenting, and intracranial angioplasty are typically recorded in the procedural description and intervention details.

Variables were selected based on two criteria: (1) consistent documentation across institutions, and (2) relevance to standardized clinical metrics. A broader set of candidate variables—such as time metrics (e.g., puncture-to-recanalization time), number of thrombectomy passes, medication dosages, and device names, were excluded due to inconsistent reporting formats and frequent documentation outside of free-text radiology reports (e.g., in structured EMR fields). These selected variables also capture the core diagnostic, interventional, and outcome metrics required for comprehensive stroke center certification.

Although these variables are routinely addressed in neurointerventional procedure reports, their phrasing and level of explicit documentation may vary substantially across cases and institutions. Accordingly, the current variable set was deliberately restricted to items that are reliably and consistently documented across sites, enabling a standardized large-scale evaluation of LLM-based extraction while maintaining clinical and operational relevance.

### Prompting Strategies for Clinical Variable Extraction

The LLMs extract structured clinical variables by reading full-text reports and selecting the most appropriate answer to each predefined prompt, based on the contextual evidence within the narrative. To enable fair and consistent evaluation across all models, we designed a standardized prompting framework for clinical variable extraction. Two distinct prompting strategies were employed: Quick Response and Chain-of-Thought (CoT) prompting.

Quick Response prompting is an instruction-driven strategy designed for efficient, high-throughput extraction of structured variables. As illustrated in Figure 3a, each target variable is framed as a discrete multiple-choice question with predefined answer choices reflecting clinically meaningful categories (e.g., presence or absence of an intervention). The LLM is provided with the full report as context and instructed to output only the final answer index (e.g., 0, 1, or None) without performing explicit reasoning. The complete set of prompt templates and output formats for all variables is provided in Table S.3 of Supplemental Data.

In contrast, Chain-of-Thought prompting introduces a lightweight reasoning scaffold to guide the model through stepwise inference before classification in Figure 3b. While using the same question-answer structure as Quick Response prompting, CoT prompts explicitly instruct the model to identify relevant information, apply clinical logic, and explain its decision prior to final output. This strategy enhances interpretability and improves performance on context-dependent or ambiguously phrased variables.

We compare these two prompting modes that offer complementary strengths: Quick Response prompting prioritizes rapid and scalable inference with minimal computational overhead, whereas CoT prompting explicitly elicits intermediate reasoning steps to support extraction of variables that may require contextual interpretation. Detailed methodology and comparisons are provided in the Supplemental Materials.

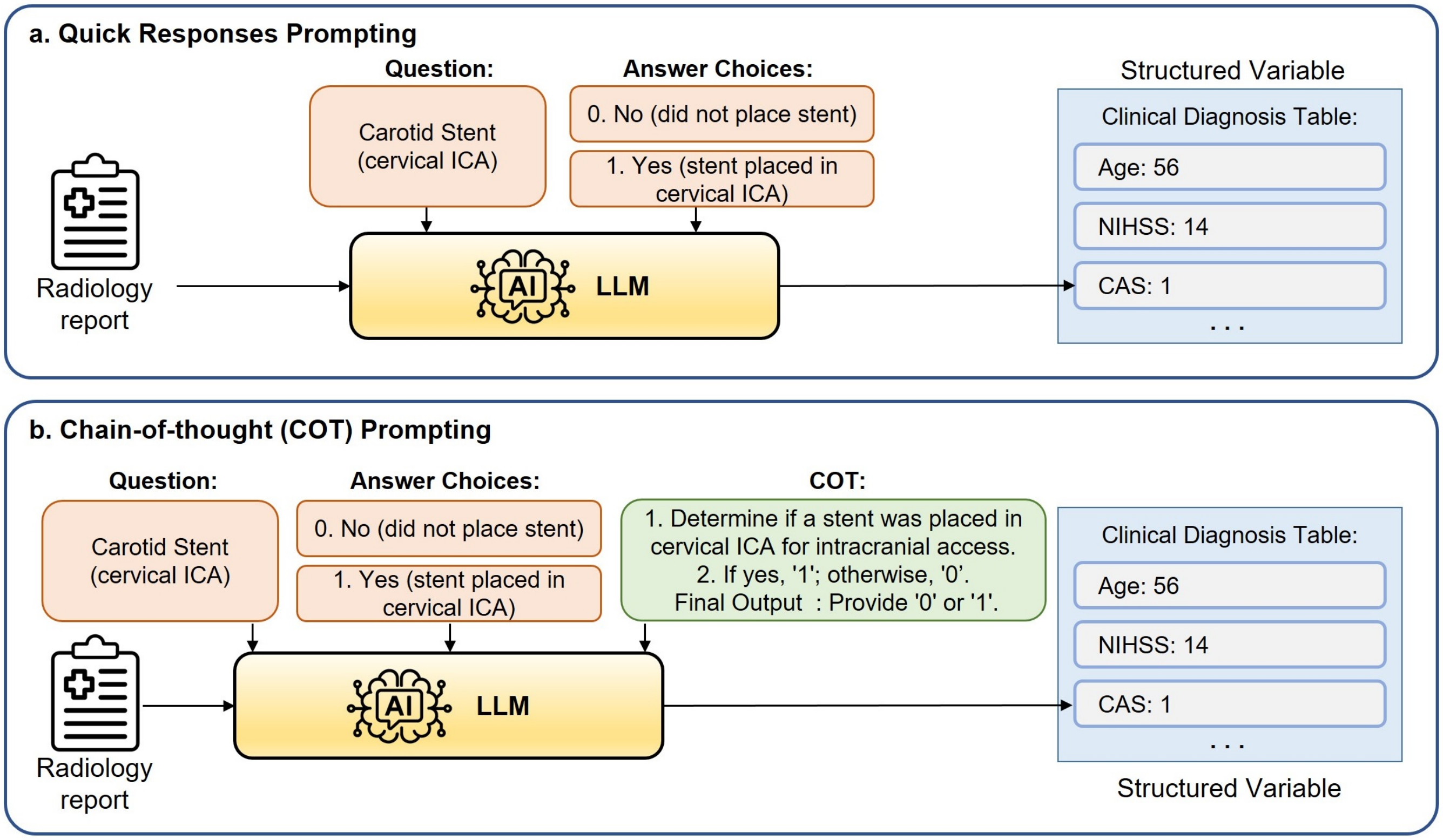

Figure 3. Comparison of prompting strategies for LLM-based clinical variable extraction. Top: Quick Response Prompting provides fixed output formats based on keyword matching. Bottom: Chain-of-Thought (CoT) Prompting introduces reasoning steps to guide the LLM in identifying context-dependent information, improving accuracy in procedural classification. Extracted outputs populate structured clinical tables for downstream analysis.

For a comprehensive evaluation, we assessed the performance of 22 instruction-tuned LLMs on both prompting strategies. These models represent seven major architecture families, including LLaMA[17], Mistral[18], Gemma[19], Phi[20], Qwen[21], HuatuoGPT[22], and OpenBioLLM[23], and encompass both general-purpose models and biomedical domain–adapted variants. All models are open-source and support local deployment, allowing secure inference within institutional computing environments.

### Statistical Analysis and Reference Standard

Model performance in structured variable extraction was evaluated using standard classification metrics. Accuracy was defined as the proportion of categorical labels correctly predicted by each LLM relative to the reference standard. For both binary and multiclass variables, predictions were considered correct only if they exactly matched the corresponding reference value. Sensitivity and specificity were used to characterize the model's ability to correctly identify the presence and absence of a given clinical variable, respectively. Precision quantified the reliability of positive predictions, and the F1 score summarized the balance between precision and sensitivity. Ninety-five percent confidence intervals were calculated to quantify statistical uncertainty.

The reference standard for this cross-model evaluation was based on non-expert annotations, defined as baseline labels generated by hospital personnel such as registry coordinators and data abstractors following training on the specific data values by physician staff. These annotators had multiple years of experience, worked under the supervision of neurointerventional faculty, received standardized training by expert faculty, and participated in institutional registry and quality improvement efforts. A majority of the source data have been reviewed by physician staff in separate research projects to ensure consistency. Table S.1 of Supplemental Data provides patient demographics and the distributions of eight key variables as annotated by non-expert staffs.

These non-expert annotators routinely abstracted data from both narrative procedure reports and structured clinical tables. The tables often contained supplemental procedural details not explicitly documented in the free-text reports, providing additional context for accurate annotation. As a result, the inherent incompleteness of real-world reports introduced two types of missingness that posed challenges for model validation: (1) cases where an annotated label was available but the relevant information was absent in the report, and (2) cases where the report contained extractable information but no reference annotation was available in the label table. These bidirectional gaps introduced uncertainty in evaluating prediction correctness.

To mitigate the impact of text–label inconsistencies, we incorporated a model-derived surrogate reference to assess whether sufficient textual evidence existed to support clinical variable extraction. Specifically, predictions from the strongest-performing model (LLaMA3.3-70B[17] with CoT prompting) were used as a proxy indicator of extractability for each report–variable pair. As shown in the left-most panel of Figure 4, the bars represent the distribution of surrogate predictions, including both correct (green/blue) and incorrect (red). The unfilled portions indicate report–variable pairs with no valid prediction, reflecting text–label mismatches and missingness as described above.

Accordingly, only instances in which the LLaMA3.3-70B with CoT prompting model produced a valid prediction with a corresponding reference label were included in the evaluation of other models. This restriction ensured that comparisons were limited to report–variable pairs with likely textual evidence, thereby reducing bias introduced by label–report mismatches.

To assess the impact of model architecture, scale, and reasoning strategy on extraction performance, we stratified accuracy by model family, parameter size, and prompting methods (Quick Response vs. Chain-of-Thought). All LLMs, including LLaMA3.3-70B, were deployed locally via the Ollama [24] platform, with each running independently on an Nvidia L40s GPU which is setup in our AI computing server within our institutional firewall to secure the data and computation.

### Expert Adjudication of Discordant between AI And Non-expert staff

In cases where AI predictions and non-expert staff annotations were discordant, expert adjudication was conducted to determine the more accurate label. From the pool of conflicting cases, 20 cases were randomly sampled for each of the eight clinical key variables. This evaluation focused on variables that required clinical interpretation or contextual reasoning, ensuring a balanced and representative evaluation of model and human performance.

Each case was independently reviewed using a previously validated web-based platform[25], where expert raters analyzed de-identified clinical reports and completed structured multiple-choice questions. The survey design, answer choices, and output formats were standardized across raters and aligned with the definitions outlined in Table S.3 of Supplemental Data. Raters were instructed to extract the most appropriate label based on the content of each report and the corresponding variable schema.

The expert evaluation panel comprised two senior stroke physicians and four medical students at various stages of their training. Both physicians had completed residency and neurointerventional fellowship training and had approximately six years of independent clinical practice following fellowship at the time of the study. All medical students had completed at least three months of supervised clinical and research training in stroke procedural outcomes.

To consolidate the expert judgments, we applied a weighted majority voting strategy that accommodates multi-class classification and accounts for rater expertise. Specifically, each rater's vote was assigned a predefined weight reflecting their clinical experience: 0.7 for each of the two senior stroke physicians, 0.4 for the advanced (fourth-year) medical student, 0.2 for the intermediate (second-year) medical student, and 0.1 for the two junior (first-year) medical students. This approach yielded a single consensus label per case, which served as the adjudicated ground truth to evaluate discrepancies between AI and non-expert staff predictions.

# Results

## Prompting Strategy Comparison

The LLaMA3.3-70B[17] model demonstrated robust and consistent performance across two prompting strategies when extracting most clinical variables. As shown in Figure 4, CoT prompting yielded higher accuracy for variables requiring contextual or stepwise inference (e.g., site of occlusion), while performance differences were minimal for directly stated procedural fields. High accuracy (>0.94) was observed for procedural variables including carotid stenting, carotid angioplasty, intracranial stenting, intracranial angioplasty, and IV TPA, under both prompting strategies. Correct prediction ratios for these variables consistently exceeded 0.95.

Variables such as NIHSS and site of occlusion had moderately lower accuracy (0.85–0.87) and higher rates of incorrect or null predictions, likely due to their reliance on indirect narrative content or inconsistent documentation. Structured variables with direct phrasing were extracted with high reliability, while fields that required inference from less standardized language showed decreased performance. Table S.4 of Supplemental Data illustrates how the CoT prompting enables the model to perform stepwise clinical reasoning and align its free-text interpretations with the predefined structured variable schema.

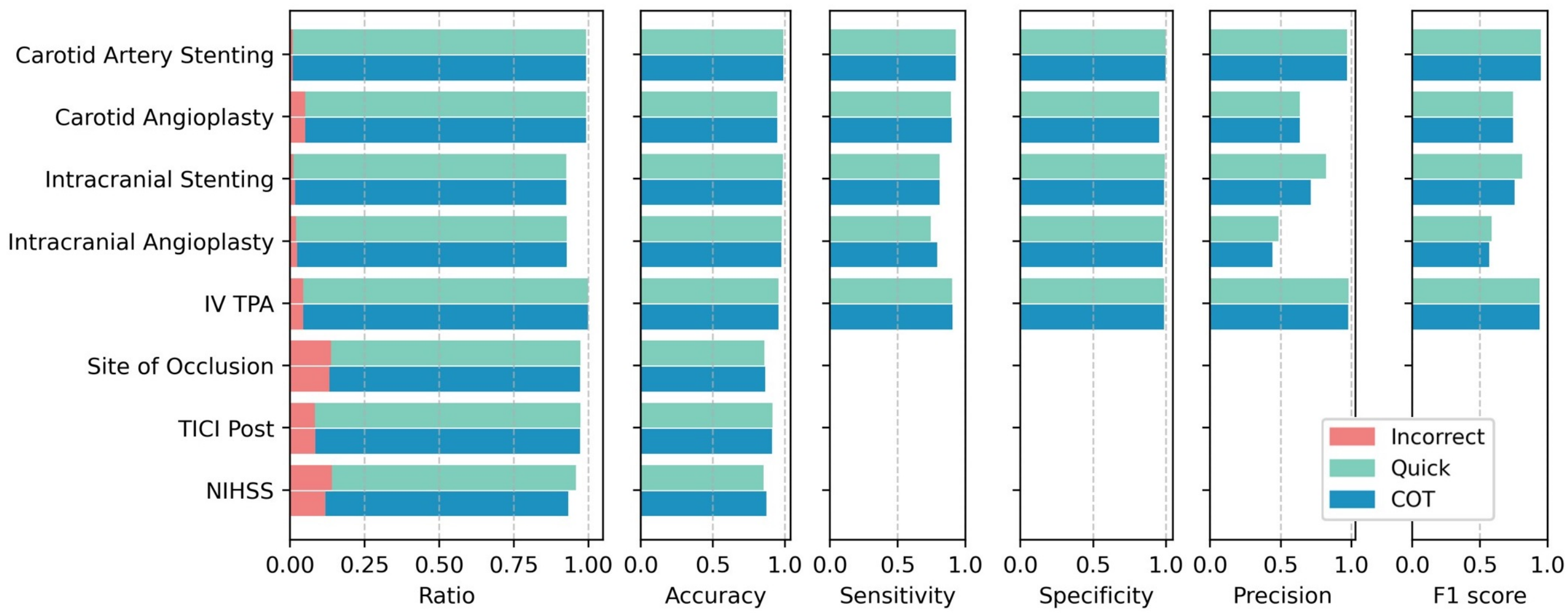

Figure 4. Comparative extraction performance of the LLaMA3.3-70B model and non-expert staff annotations across 8 clinical variables. The left-most panel shows prediction distribution for the LLaMA3.3-70B model using Quick Response (green) and Chain-of-Thought (blue) prompting, with correct (green/blue), incorrect (red), and invalid (void) predictions. The remaining panels summarize accuracy, sensitivity, specificity, precision, and F1 score for LLaMA3.3-70B predictions, evaluated against the non-expert staff-provided annotations as reference baselines.

## Model Size and Architecture Comparison

As shown in Figure 5, twenty-two LLMs were grouped into four categories based on parameter size: Tiny (<7B), Small (7–12B), Medium (12–30B), and Large (≥30B). Across all sizes, CoT prompting yielded better or equal performance compared to Quick Response prompting. Among large models, LLaMA3.3-70B achieved near-ceiling accuracy on all variables (e.g., carotid stenting: 0.991, TICI post: 0.915, IV tPA: 0.955, NIHSS: 0.873). Other large modelss such as Qwen2.5-72B, Qwen3-30B, and Qwen-Q[21] also maintained high accuracy (typically >0.90) with both prompting strategies.

In the medium group, Gemma3-27B[19] and MedGemma3-27B[26] performed strongly on procedural variables (intracranial stenting: 0.975–0.987; IV tPA: 0.955). Within the small-scale models, Qwen3-8B, HuatuoGPT-o1-8B[22], and OpenBio-LLaMA3-8B[23] showed robust performance. HuatuoGPT-o1-8B excelled on intervention and demographic variables (e.g., IV tPA: 0.955 under CoT). Biomedical-tuned models (e.g., HuatuoGPT, OpenBioLLM) consistently outperformed their corresponding general-purpose models.

The tiny models showed more variable and generally lower performance, especially on nuanced fields. While Gemma3-4B and Qwen2.5-3B underperformed on TICI post, site of occlusion, and NIHSS (often <0.70).

Across model groups, CoT prompting showed larger performance gains for inference-dependent variables, such as site of occlusion, compared with directly documented fields. For example, Qwen3-8B improved from 0.837 (Quick) to 0.853 (CoT) on site of occlusion, and HuatuoGPT-o1-8B improved from 0.616 to 0.747 on NIHSS.

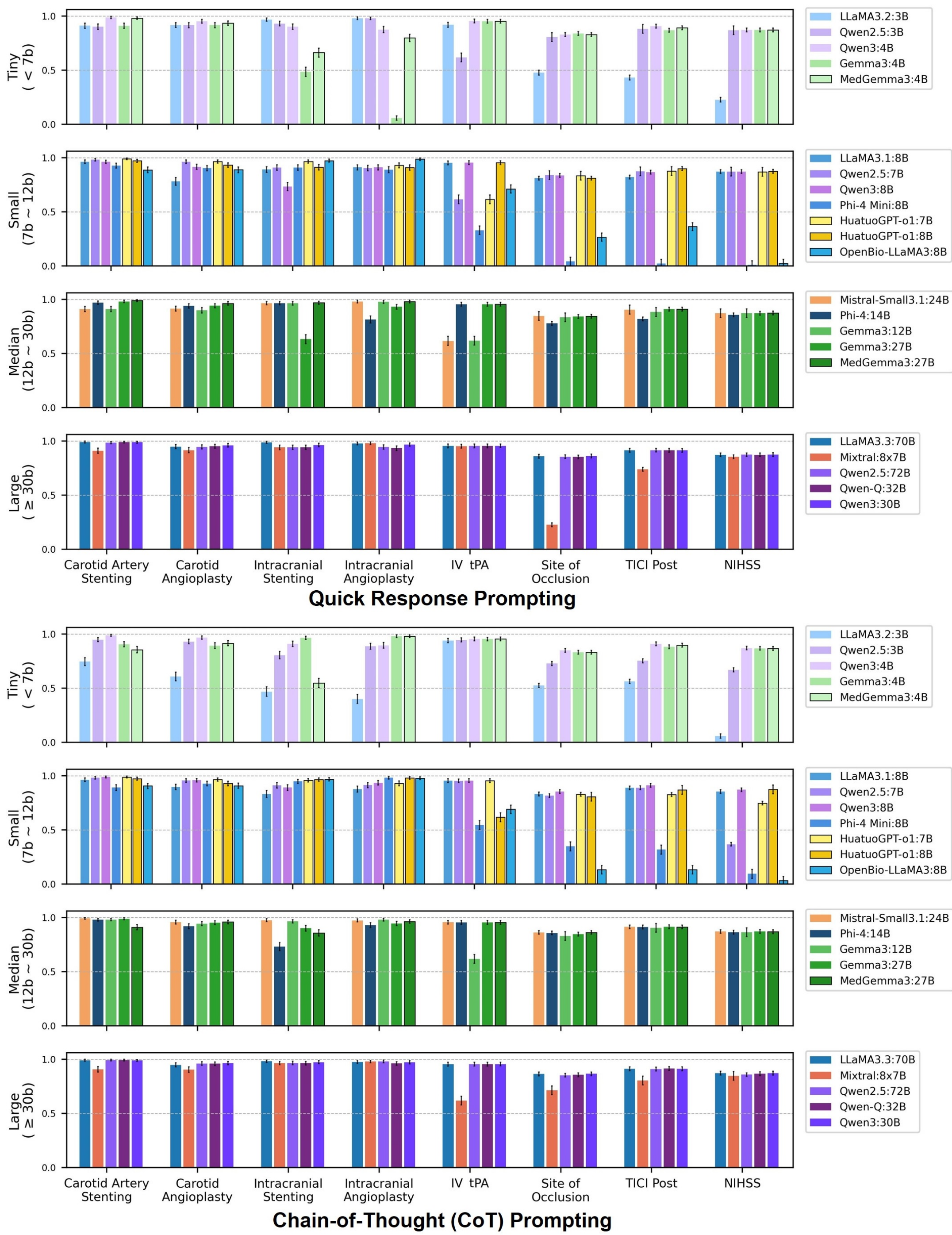

Figure 5. Accuracy comparison of LLMs on the clinical variables extraction using Quick Response Prompting (top) and Chain-of-Thought (CoT) Prompting (bottom). Models are grouped by size: Tiny (<7B), Small (7–12B), Medium (12–30B), and Large (≥30B). The colors with bounding box represent the medical fine-tuning LLMs. Error bars with caps indicate the 95% confidence intervals.

### Inference Time vs. Accuracy

Quantitative trade-off analysis demonstrated that inference time scaled with model size and prompting complexity. Under Quick Response prompting, as illustrated in Figure 6, large models like LLaMA3.3-70B and Qwen2.5-72B achieved top accuracy (0.938–0.948) with relatively short latencies (<10s). Qwen3-30B and Qwen-Q had longer latencies (~45–60s), due to internal reasoning paradigms even under the Quick prompts.

Medium models such as MedGemma3-27B, Gemma3-27B, and Mistral-Small3.1-24B showed strong accuracy (≥0.90) with moderate inference times (10–45s). Small models like Qwen3-8B and HuatuoGPT-o1-8B reached high accuracy (>0.89) with sub-10-second latency. In contrast, OpenBio-LLaMA3-8B and Phi-4 Mini:8B had lower accuracy (≤0.60) despite fast inference.

Under CoT prompting, latency increased across all models. Top-performing models—Qwen3-30B, Mistral-Small3.1-24B, and LLaMA3.3-70B—achieved accuracy of 0.948–0.949, but with latency approaching 190s. Qwen2.5-72B and MedGemma3-27B remained highly accurate (>0.92), though inference time exceeded 130s.

Medium models like Phi-4-14B and Gemma3-12B maintained accuracy between 0.91 and 0.93, with inference time between 17 and 52 seconds. The smallest models, such as LLaMA3.2-3B and Qwen2.5-3B, had the fastest inference (16–29s), but accuracy dropped below 0.85.

Notably, CoT prompting improved accuracy by 0.13 for Gemma3-4B and 0.03 for Gemma3-27B. These findings indicate that CoT boosts accuracy, especially in models of intermediate size, at the cost of inference latency.

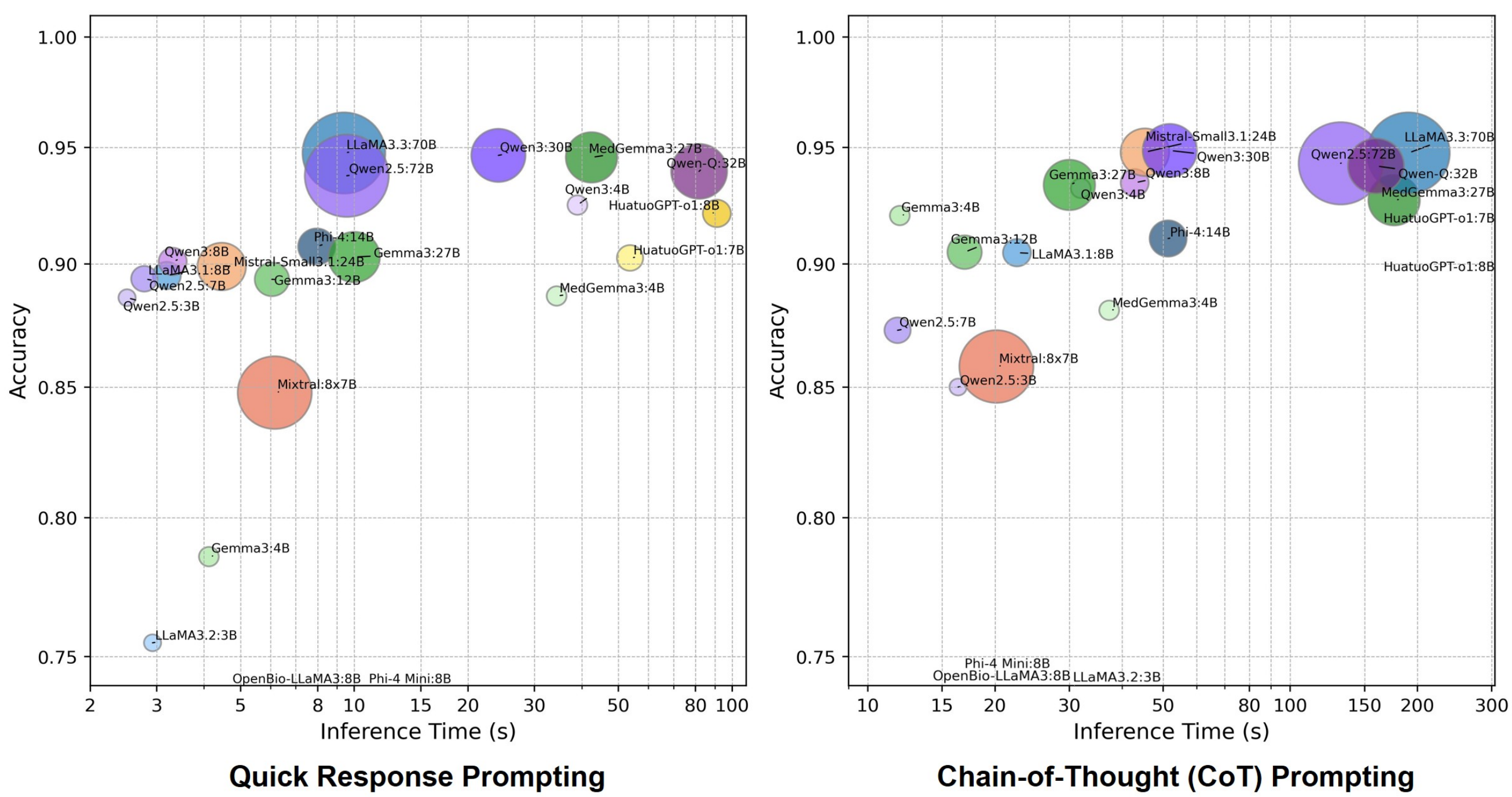

Figure 6. Inference time vs. accuracy trade-off across LLMs using Quick Response Prompting (left) and Chain-of-Thought (CoT) Prompting (right). Each bubble represents a model, with its position determined by case-averaged inference time (x-axis) and extraction accuracy (y-axis). Bubble size reflects model parameter count, and color indicates model family.

### AI vs. Non-expert Staff Accuracy

When AI predictions (LLaMA3.3-70B with CoT) and non-expert annotations disagreed, expert consensus was used to assess correctness. In Figure 7a, AI consistently outperformed non-expert staff on most variables except for Intracranial Angioplasty. The AI reached 100% accuracy on IV tPA, compared to 0% by staff. On NIHSS, AI reached 0.95, while staff had only 0.05. Similarly, TICI Post accuracy was 0.90 for AI vs. 0.05 for staff.

Procedural variables showed similar trends: carotid stenting (AI: 0.80 vs. staff: 0.20), carotid angioplasty (AI: 0.70 vs. staff: 0.30). On intracranial stenting, both groups performed equally (0.50), while intracranial angioplasty was an exception, where non-expert staff achieved 0.95 versus AI's 0.05.

### Rater Accuracy across Clinical Variables

Six adjudicators (two senior physicians and four medical students) participated in an adjudication of discordant between AI and non-expert, independently reviewed and consolidated into a consensus label through a weighted majority voting. In Figure 7b, the senior physicians and the advanced medical student consistently achieved high accuracy (≥0.95) across all variables.

On structured fields like IV TPA and NIHSS, raters scored 1.00. For procedural variables, accuracy varied by experience: e.g., for carotid stenting, physicians and upper-year students achieved ≥0.95, while junior medical student B dropped to 0.65. Similarly, intracranial angioplasty accuracy ranged from 0.95 (physicians) to 0.50 (junior student). Despite this, even junior students performed reasonably well (≥0.85) on fields like TICI Post and Site of Occlusion, suggesting clear documentation supports consistent annotation.

### Agreement Between Expert Raters and AI vs. Non-expert

Cohen's kappa[27] was used to assess agreement between each rater and AI or staff annotations. As shown in Figure 7c–d, agreement between AI predictions and expert annotations consistently exceeded agreement with non-expert staff. For structured variables (e.g., IV TPA, TICI Post, NIHSS), all raters, even junior ones, achieved kappa scores ≥0.82 with AI. In contrast, agreement with non-expert staff was low or zero (e.g., NIHSS = 0.06; IV TPA = 0).

For site of occlusion, agreement with AI was moderate (0.62–0.75) but poor with staff. On procedural fields like carotid PTA, agreement with AI remained modest (~0.30–0.60), and often negative with staff (e.g., –0.48 for Student B).

Intracranial angioplasty was an outlier where AI had zero agreement with all raters, while staff had moderate-to-high agreement (up to 1.00), indicating that these reports exhibit systematic interpretation differences.

Inter-rater heatmaps (in Supplemental Data) confirmed these findings: expert adjudicators approached perfect agreement (0.90-1.00) on structured variables but showed lower consistency on procedural fields, particularly among junior raters.

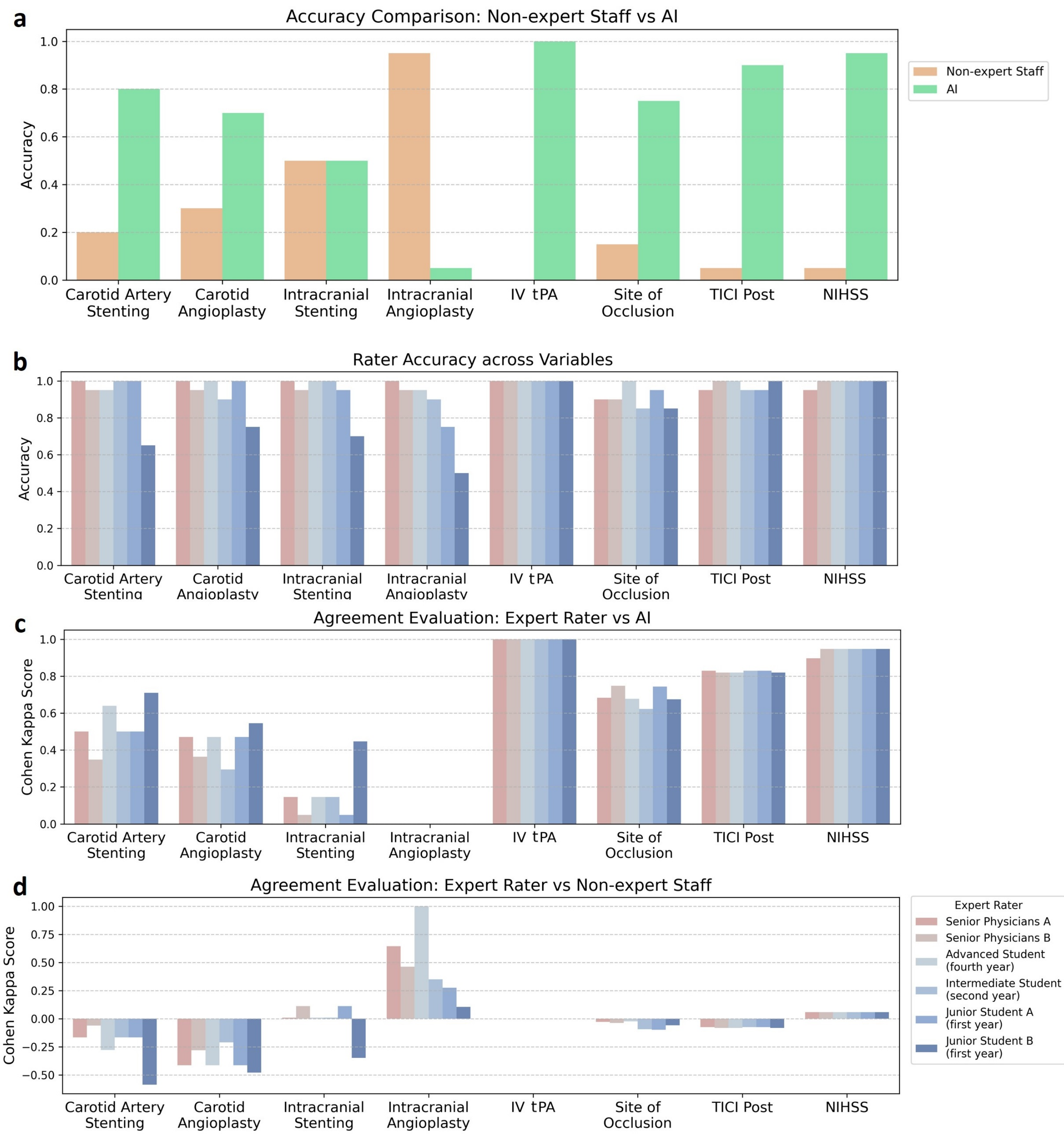

Figure 7. Expert adjudication analysis on inconsistent AI and non-expert staff cases. (a) Accuracy comparison between AI and non-expert staff predictions across eight clinical variables, using expert-voted labels as ground truth. (b) Individual rater accuracy relative to the expert consensus. (c) Agreement between each rater and the AI model, measured by Cohen's kappa. (d) Agreement between each rater and the non-expert staff annotations.

## Discussion

This large-scale benchmarking study evaluated 22 open-source LLMs across architectures, sizes, and prompting strategies for structured variable extraction from stroke procedure reports. High-performing models such as LLaMA3.3-70B[17] and Qwen3-30B[21] achieved near-expert accuracy in identifying key diagnostic and procedural variables, demonstrating their potential as scalable AI assistants for clinical data abstraction. CoT prompting enhanced accuracy, particularly for reasoning-dependent variables, by guiding stepwise clinical interpretation aligned with structured schemas. However, it also increased inference latency, indicating a trade-off between reasoning depth and efficiency. Some reasoning-tuned models showed limited adaptability to required response, often generating unnecessary reasoning chains that increased latency and token use, reduce efficiency for concise classification. Domain-adapted biomedical models showed heterogeneous gains, MedGemma3-27B[26] performed competitively, while others (e.g., OpenBioLLM[23]) offered limited improvement, highlighting that corpus quality and alignment strategy are as critical as domain exposure.

The relatively lower performance in identifying intracranial angioplasty likely reflects both linguistic and data-driven challenges. This procedure was described inconsistently across reports, with terminology varying by author and training level with expressions such as balloon angioplasty, intracranial angioplasty, balloon remodeling, or treatment with a named, specific balloon device. Such heterogeneity in phrasing limited the LLM's ability to consistently map narrative text to a standardized label. In addition, intracranial angioplasty represents a less frequently performed intervention compared to stenting or thrombectomy, resulting in a smaller sample size and fewer training instances. The combination of limited data and variable terminology likely contributed to higher classification variability and reduced accuracy for this field. Future iterations of the model may benefit from synonym-aware prompting, terminology normalization, or integration with structured device metadata to improve robustness on low-frequency procedural variables.

Expert adjudication focused on cases where AI predictions and non-expert annotations disagreed, representing the most challenging reports requiring contextual reasoning and clinical interpretation. The analysis revealed a clear link between clinical experience and annotation accuracy: senior stroke physicians and a 4th-year medical student achieved high accuracy ($\geq$0.95) across variables, while first-year students performed well on structured fields (e.g., IV tPA, NIHSS) but struggled with complex procedures such as intracranial angioplasty or carotid stenting. Within this difficult subset, the LLM-based system achieved performance largely comparable to non-expert raters and demonstrated stronger agreement with experts, particularly for well-defined variables. These findings support a hybrid annotation framework, reserving complex variables for experienced clinicians while delegating standardized fields to non-experts or AI systems.

In conclusion, these results highlight the dual benefit of high-performing LLMs: not only can they automate structured data extraction with strong and reproducible accuracy, but they also exhibit consistency comparable to expert clinicians for clearly defined variables. However, variable-specific ambiguity, which arises from heterogeneous documentation or procedural interpretation, remains a key challenge for both AI and junior annotators. By transforming free-text reports into structured clinical tables, these models have the potential to streamline clinician workflows, enhance data completeness, and reduce the burden of manual retrospective chart review, enabling more efficient quality monitoring and research-driven registry maintenance. These findings underscore the importance of maintaining expert oversight and implementing targeted quality control mechanisms when scaling AI-assisted chart abstraction in clinical research or real-world deployment.

Despite the large-scale evaluation across three hospitals, this study has several limitations. First, the reference standard relied primarily on non-expert annotations derived from routine registry and medical chart abstraction workflows. Although these annotations reflect real-world clinical practice and were generated by experienced personnel under standardized training and supervision, they may introduce labeling variability compared with expert-reviewed ground truth, particularly for variables requiring nuanced clinical interpretation.

Second, to mitigate inconsistencies between free-text reports and reference labels, we used the strongest-performing model as a surrogate filter to identify evaluable report–variable pairs. While this strategy reduced bias arising from documentation incompleteness, it may introduce selection bias by aligning the evaluation subset with the behavior of the reference model.

Third, expert adjudication was conducted on a relatively small subset of cases selected for discordance between AI predictions and non-expert annotations. Although this design focused expert review on the most challenging and clinically ambiguous cases, the adjudicated subset may not fully represent the distribution of all reports in the cohort, and conclusions drawn from this analysis should be interpreted accordingly.

Finally, the generalizability of the proposed LLM-based framework is constrained by the institutional and linguistic scope of the dataset. The evaluation was conducted within a single health system, and reporting templates, documentation styles, and terminology conventions may differ across institutions, potentially affecting extraction performance in external settings. In addition, although Quick Response and Chain-of-Thought prompting represent efficiency- and reasoning-oriented paradigms, alternative strategies such as retrieval-augmented or few-shot prompting may further improve extraction accuracy or interpretability.

## Conclusions:

In this large-scale, multi-institutional study, we systematically evaluated the ability of instruction-tuned LLMs to automate structured clinical variable extraction from neurointerventional procedure reports in patients with acute ischemic stroke. Our findings demonstrate that state-of-the-art LLMs, particularly LLaMA3.3-70B with chain-of-thought prompting, can accurately and scalably extract a range of diagnostic and procedural variables from free-text reports, competing non-expert staff and matching the accuracy of junior medical trainees on inference-dependent fields. Notably, the use of chain-of-thought prompting substantially improved model performance on variables requiring contextual interpretation, highlighting the importance of both model architecture and prompting strategy in clinical natural language processing applications. Expert adjudication further established that AI predictions aligned more closely with clinical interpretations than those of non-expert annotators, particularly for structured outcome measures. These results support the feasibility of integrating locally deployed LLMs into existing clinical research pipelines and automated registry curation workflows, potentially streamlining retrospective data abstraction and improving the consistency and efficiency of large-scale quality monitoring in stroke care. Future research should focus on broader external validation and the development of hybrid AI–human workflows to further enhance accuracy and clinical adoption.

### Data Availability Statement

Due to the presence of protected health information (PHI) and compliance with HIPAA and institutional review board (IRB) requirements, the datasets generated and/or analyzed during the current study are not publicly available. De-identified data may be made available upon reasonable request to the corresponding author and with IRB approval.

### Funding

This study received no external funding and was conducted using intramural resources only.

## Supplemental material

### Detailed Methodology

To enable standardized and reproducible evaluation across all large language models (LLMs), we designed a unified prompting framework tailored for structured clinical variable extraction from radiology and neurointerventional procedure reports. The framework consists of two complementary prompting strategies: Quick Response Prompting and Chain-of-Thought (CoT) Reasoning. Both approaches operate under a constrained multiple-choice classification schema, but differ in their reasoning complexity and intended use cases. Full prompt templates and response examples for all variables are provided in Table S.3.

### Quick Response Prompting

Quick Response Prompting is a lightweight and instruction-driven strategy designed to support rapid, high-throughput inference of clinical variables. Each target variable is framed as a discrete multiple-choice question, with clearly defined answer options that directly map to clinically interpretable categories (e.g., binary or ternary: Yes, No, Unknown). This format ensures consistency across variables and facilitates automated downstream processing.

In this setting, the model receives:

The complete radiology or procedure report as the input context

A structured question about the target variable

A predefined set of answer options

An explicit instruction to return the final answer only, without explanation or justification

The instruction explicitly discourages reasoning, encouraging the model to rely on its internal pattern recognition and learned associations. This format is particularly suitable for well-structured or explicitly documented variables, such as stent placement or TICI scores.

A representative prompt example for the variable Carotid Stent (cervical ICA) is shown below:

> "Context: [full radiology report]
>
> Question: Was a carotid stent placed in the cervical internal carotid artery (ICA)?
>
> Answer Choices:
>
> – '0' = No (did not place stent)
>
> – '1' = Yes (stent placed in cervical ICA)
>
> – 'None' = Unknown (status unclear)
>
> Final Output: Provide 0, 1 or None.
>
> Please give me a quick, instinctual response to this question without detailed reasoning. Only provide the answer, no thinking or elaboration allowed.
>
> At the end of the output, using this as a conclusion format 'The final answer is: [the number answer]'. If you can't find any relevant content, using this as a conclusion format 'The final answer is: None'. Using plain text output."

This method maximizes inference efficiency and reduces the risk of hallucinated rationales, while preserving output format uniformity for variable extraction at scale.

### Chain-of-Thought Reasoning Prompting

Chain-of-Thought (CoT) Prompting is a reasoning-augmented strategy designed to enhance model interpretability and performance for context-dependent or linguistically ambiguous variables. It encourages the model to simulate step-by-step clinical reasoning, enabling it to better handle indirect cues, latent inferences, and complex syntax often present in narrative medical documentation.

The overall structure mirrors that of Quick Response Prompting, with the addition of:

A reasoning scaffold (Chain-of-Thought section) before final output

Instructions for the model to identify relevant report content, apply clinical logic, and then derive the correct label

The answer format remains consistent (i.e., 0, 1, or None), ensuring compatibility with automated scoring pipelines. The prompt encourages explicit thought articulation, helping to disentangle the model's reasoning path from its final decision.

A representative CoT prompt for the same variable (Carotid Stent) is:

"Context: [full radiology report]

Question: Was a carotid stent placed in the cervical internal carotid artery (ICA)?

Answer Choices:

– '0' = No (did not place stent)

– '1' = Yes (stent placed in cervical ICA)

– 'None' = Unknown (status unclear)

**Chain-of-Thought:**

**1. Review the report for mention of a stent placement procedure in the cervical ICA region.**

**2. Determine whether the placement was performed or planned.**

**3. If a stent was placed, return '1'.**

**4. If the report indicates no stenting, return '0'.**

**5. If the report is ambiguous or lacks information, return 'None'.**

Final Output: Provide 0, 1 or None.

**Please output your complete thought process before providing the final answer.**

At the end of the output, using this as a conclusion format 'The final answer is: [the number answer]'. If you can't find any relevant content, using this as a conclusion format 'The final answer is: None'. Using plain text output."

The bold text indicates reasoning prompts that differ from quick-response prompting. By incorporating CoT reasoning, the model is not only prompted to respond but also to internally justify its decision. This approach enables the handling of tasks involving nuanced language, implicit references, and ambiguous phrasing—features commonly encountered in clinical documentation. CoT prompting thus provides a principled structure for eliciting stepwise reasoning in clinical information extraction.

### LLM Selection and Benchmarking Across Model Families and Sizes

In this study, we evaluated a diverse set of open-source LLMs. Models were grouped by series, and multiple sizes within each series were tested to assess performance scalability. All models were evaluated in their instruction-tuned formats using either quantized or FP16 precision settings to ensure efficient and consistent inference.

The LLaMA [1] series, developed by Meta, is widely regarded for its strong generalization and competitive performance across multiple benchmarks. We included versions ranging from lightweight (3B) to high-capacity models (70B), with the LLaMA3.3–70B model selected as our reference baseline due to its state-of-the-art performance in both reasoning and extraction tasks.

The Mistral [2] series is known for its efficiency and strong performance in long-context understanding and multilingual applications. We included the 24B variant of Mistral-small3.1, which has demonstrated robust instruction-following capabilities.

The Gemma [3] series, released by Google, emphasizes high-quality instruction tuning and alignment with downstream prompts. We evaluated 4B, 12B, and 27B models from this series to explore trade-offs between model scale and extraction accuracy in the clinical domain.

The Phi [4] series, developed by Microsoft, is designed for efficient reasoning and compact model performance. The 14B Phi-4 model was included to assess its ability to perform structured information extraction under constrained computational settings.

The Qwen [5] series, developed by Alibaba, is optimized for multilingual reasoning and general-purpose instruction tasks. We tested models at 3B, 7B, 32B, and 72B scales, enabling detailed comparison across capacities. The 32B reasoning-optimized Qwen-Q–32B was additionally examined for its performance in CoT reasoning scenarios.

To evaluate the impact of biomedical domain adaptation, we included models from HuatuoGPT [6] and OpenBioLLM [7]. These models are fine-tuned on medical corpora and clinical data, with the goal of enhancing performance in healthcare-specific tasks. Specifically, we assessed HuatuoGPT [6] variants based on both LLaMA3.1-8B (HuatuoGPT-o1-8B) and Qwen2.5-7B (HuatuoGPT-o1-7B) backbones, as well as OpenBioLLM [7] based on LLaMA3.1-8B. In addition, we included MedGemma3[8], a medically adapted variant of the Gemma series, designed to incorporate domain-specific knowledge into the Gemma backbone (4B and 27B).

Table S.1. Patient demographics (e.g., sex and age) and distribution of the eight clinically relevant variables from neurointerventional procedure reports.

| PARAMETER | CATEGORY | Patients (n=2416) |
|---|---|---|
| Sex | Male | 1194 (49.42%) |
| Sex | Female | 1222 (50.58%) |
| Age (years) | ≤ 65 | 804 (33.28%) |
| Age (years) | 65 – 80 | 816 (33.77%) |
| Age (years) | ≥ 80 | 796 (32.95%) |
| Carotid Artery Stenting | No | 2180 (90.23%) |
| Carotid Artery Stenting | Yes | 199 (8.24%) |
| Carotid Artery Stenting | N/A | 37 (1.53%) |
| Carotid Angioplasty | No | 2194 (90.81%) |
| Carotid Angioplasty | Yes | 187 (7.74%) |
| Carotid Angioplasty | N/A | 35 (1.45%) |
| Intracranial Stenting | No | 2144 (88.74%) |
| Intracranial Stenting | Yes | 73 (3.02%) |
| Intracranial Stenting | N/A | 199 (8.24%) |
| Intracranial Angioplasty | No | 2181 (90.27%) |
| Intracranial Angioplasty | Yes | 43 (1.78%) |
| Intracranial Angioplasty | N/A | 192 (7.95%) |
| IV Thrombolytics | No | 1543 (63.87%) |
| IV Thrombolytics | Yes | 868 (35.93%) |
| IV Thrombolytics | N/A | 5 (0.21%) |

Table S.1. Patient demographics (e.g., sex and age) and distribution of the eight clinically relevant variables from neurointerventional procedure reports. (continued)

| PARAMETER | CATEGORY | Patients (n=2416) |
|---|---|---|
| Site of Occlusion | 1. ICA (Internal Carotid Artery) | 383 (15.85%) |
| Site of Occlusion | 2. M1 (1st segment of MCA) | 1335 (55.26%) |
| Site of Occlusion | 3. M2 (2nd segment of MCA) | 399 (16.51%) |
| Site of Occlusion | 4. Basilar artery | 143 (5.92%) |
| Site of Occlusion | 5. Other vessel | 77 (3.19%) |
| Site of Occlusion | N/A | 79 (3.27%) |
| TICI Post | 0. No perfusion | 99 (4.10%) |
| TICI Post | 1. Minimal perfusion | 15 (0.62%) |
| TICI Post | 4. 2a (partial perfusion <50%) | 98 (4.06%) |
| TICI Post | 5. 2b (partial perfusion >50%) | 591 (24.46%) |
| TICI Post | 6. 2c (near-complete perfusion except small distal branches) | 538 (22.27%) |
| TICI Post | 3. 3 (complete perfusion) | 990 (40.98%) |
| TICI Post | N/A | 85 (3.52%) |
| NIHSS | 0 | 13 (0.54%) |
| NIHSS | 1 - 4 | 161 (6.66%) |
| NIHSS | 5 - 9 | 376 (15.56%) |
| NIHSS | ≥10 | 1806 (74.75%) |
| NIHSS | N/A | 60 (2.48%) |

Table S.2. An anonymized sample interventional radiology report for acute stroke intervention

HISTORY: VIR service was consulted for Acute Stroke intervention.

Pre-procedure Diagnosis: Intracranial embolism of inferior division of the left M2

Post-procedure Diagnosis: Intracranial embolism of inferior division of the left M2

Procedure: Extirpation of intracranial embolism of inferior division of the left M2

Last known normal time: [Time Redacted].

VIR Service notified at [Time Redacted].

NIHSS: 9.

IV tPA was administered.

Groin access at [Time Redacted].

OPERATOR(S): [IR Fellow, MD] and [Attending Interventional Radiologist, MD], attending interventional radiologist who was present for and actively supervised the entire procedure.

ANESTHESIA: Local anesthesia was obtained with buffered 1% lidocaine. Intravenous conscious sedation and monitoring was provided by the attending interventional radiologist and nursing staff. Total doses of Versed and Fentanyl as per the electronic medical record.

CONTRAST: 140 mL intra-arterial Omnipaque-300

OTHER MEDICATIONS: None

ESTIMATED BLOOD LOSS: Minimal

IMPLANTABLE DEVICES/CATHETERS/SPECIMENS: None

FLUOROSCOPY TIME/ DOSE AREA PRODUCT: 10.7 minutes/378.38 mGy.

COMPLICATIONS: None.

DISPOSITION :Stable and awake on transportation to recovery room.

DESCRIPTION OF PROCEDURE:

Written and oral informed consent was obtained from the patient after explaining all the risks and benefits of the procedure as well as the alternatives.

Patient was brought to the angiography suite. Patient verification and a timeout procedure were performed as per standard hospital procedure. The groin was prepped and draped in the usual sterile fashion. All elements of maximum sterile barrier were used including mask, cap, sterile gown, gloves, large sterile sheet, hand hygiene and cutaneous antisepsis with 2% chlorhexidine. Arterial access into the right common femoral artery was obtained with a micropuncture needle in the standard fashion, and a 5 French vascular sheath was placed. The sheath was flushed and connected to continuous flush.

Selective Catheterization(s) of the [left internal carotid artery] was performed with a VER catheter.

FINDINGS:

Digital subtraction angiography was performed demonstrating occlusion of the inferior division of the left M2

INTERVENTION:

Based on angiographic findings we elected to proceed with intervention. A bolus of 3000 units IV heparin was administered. A Shuttle sheath was placed into the left internal carotid artery and subsequently a combination of 5Max ace intermediate catheter and a microcatheter was placed with access into the left middle cerebral artery. Gentle injection of contrast via the microcatheter confirmed adequate placement. The 4 mm x 30 mm TREVO stentriever was deployed across the thrombus. Gentle injection of contrast demonstrated occlusion of the inferior division of the left M2. Local aspiration with penumbra was performed and a total of one pass was made. Post-thrombectomy angiogram demonstrated TICI 2C flow. Final angiographic runs demonstrated no distal emboli.

GROIN CLOSURE: All existing catheters and wires were removed. Injection of the existing groin sheath, performed to assess for suitability for use of a closure device, demonstrated favorable anatomy. An 8French AngioSeal device was then deployed in the standard fashion and hemostasis was obtained without difficulty..

The patient was removed from the angiographic table at which point repeat physical exam demonstrated worsening right upper extremity weakness and facial droop. At this time, the patient was again placed on the angiographic suite table. The groin was again prepped and draped in the usual sterile fashion. All elements of maximum sterile barrier were used including mask, cap, sterile gown, gloves, large sterile sheet, hand hygiene and cutaneous antisepsis with 2% chlorhexidine. Arterial access into the right common femoral artery was obtained with a micropuncture needle in the standard fashion. A 5 French vascular sheath was again placed. The sheath was flushed and connected to continuous flush.

Selective catheterization of the left internal carotid artery was again performed with aVR catheter. Repeat digital subtraction angiography was performed demonstrating no evidence of emboli. There is redemonstration of TICI 2C Flow.

IMPRESSION:

1. Angiogram for acute stroke intervention demonstrates occlusion of the inferior division of the left M2 segment.
2. Post intervention, there is TICI 2C flow restored.
3. Post intervention patient demonstrated worsening upper extremity weakness (right side) and facial droop. Repeat cerebral angiography demonstrates no recurrence of occlusion with unchanged TICI 2C flow. Stat CT head was ordered.

Findings were communicated to the neurology team.

Reading Physician: [IR Fellow, MD]

Signing Doctor: [Attending Interventional Radiologist, MD] Date Signed: [Time Redacted]

Patient MRN: [Patient MRN] Accession Number: [Accession Number]

DOS: [Date Redacted] Exam: IR5699 IR ACUTE STROKE INTERVENTION

Table S.3. Standardized LLM prompt templates for clinical variable extraction with chain-of-thought reasoning sample and output formats.

| Variable | Question | Answer Choices (with clinical meaning) | Chain-of-Thought Prompt | Output Format |
|---|---|---|---|---|
| Carotid Artery Stenting | Was a carotid stent placed in the cervical internal carotid artery (ICA)? | - '0' = No (did not place stent)<br>- '1' = Yes (stent placed in cervical ICA) | 1. Determine if a stent was placed in cervical ICA for intracranial access.<br>2. If yes, '1'; otherwise, '0'." | Provide '0' or '1'. |
| Carotid Angioplasty | Was carotid angioplasty performed in the cervical segment of the internal carotid artery (ICA)? | - '0' = No (not performed)<br>- '1' = Yes (performed angioplasty of the cervical ICA) | 1. Check if cervical ICA angioplasty was performed for better intracranial access.<br>2. If yes, code '1'; if no, '0'. | Provide '0' or '1'. |
| Intracranial Stenting | Was an intracranial stent placed during the procedure? | - '0' = No (no intracranial stent placed)<br>- '1' = Yes (intracranial stent placed) | 1. Determine if intracranial stent was deployed for flow-limiting lesion.<br>2. If yes, '1'; if no, '0'. | Provide '0' or '1'. |
| Intracranial Angioplasty | Was intracranial angioplasty performed during the procedure? | - '0' = No (did not do intracranial angioplasty)<br>- '1' = Yes (angioplasty performed for intracranial lesion) | 1. Check if intracranial angioplasty was done for a flow-limiting stenosis.<br>2. If yes, '1'; if no, '0'. | Provide '0' or '1'. |
| IV Thrombolytics | Was IV thrombolysis administered? | - '0' = No (no IV tPA or TNK given)<br>- '1' = Yes (IV tPA/alteplase or TNK was given) | 1. Determine if patient received IV alteplase or tenecteplase.<br>2. If yes, code '1'; if no, '0'. | Provide '0' or '1'. |
| Site of Occlusion | What is the anatomical location of the intracranial arterial occlusion? | - '1' = ICA (Internal Carotid Artery)<br>- '2' = M1 (1st segment of MCA)<br>- '3' = M2 (2nd segment of MCA)<br>- '4' = Basilar artery<br>- '5' = Other vessel | 1. Identify which vessel is occluded from imaging or notes.<br>2. Match to the correct code. | Provide one digit from [1…5]. |
| TICI Post | What is the final modified Thrombolysis in Cerebral Infarction (TICI) score after the procedure? | - '0' = No perfusion<br>- '1' = Minimal perfusion<br>- '4' = 2a (partial perfusion <50%)<br>- '5' = 2b (partial perfusion >50%)<br>- '6' = 2c (near-complete perfusion except small distal branches)<br>- '3' = 3 (complete perfusion) | 1. Review recanalization after endovascular treatment.<br>2. Decide which TICI code applies. | Provide one code from [0,1,4,5,6,3]. |
| NIHSS | What is the Admission NIHSS (National Institutes of Health Stroke Scale)? | - Integer [0..42], where 0 = no deficit, 42 = severe stroke | 1. Identify the NIHSS prior to any therapy.<br>2. Must be 0~42. | Provide one integer in [0…42] |

Table S.4. Structured Variable Extraction from Table S.2 Sample Report of Non-Expert Annotation and LLaMA3.3-70B CoT Output. The independent inference demonstrates the model's capacity to perform stepwise clinical reasoning and align with structured schema in free-text interpretation.

| Variable | Non-Expert Annotation | LLAMA 3.3 Chain-of-Thought Output |
|---|---|---|
| Carotid Artery Stenting | 0 | To answer the question regarding the placement of a carotid stent (cervical ICA) during the procedure for the stroke patient, we need to carefully review the provided clinical report.<br>Determine if a stent was placed in cervical ICA for intracranial access: The report details the procedure for acute stroke intervention, focusing on the removal of an intracranial embolism from the inferior division of the left M2 segment. It mentions the use of various catheters and devices for accessing and treating the occlusion, including a Shuttle sheath, a 5Max ace intermediate catheter, a microcatheter, and a TREVO stentriever. However, the term "stentriever" refers to a device used for mechanical thrombectomy, not a stent left in place.<br>Review of stent placement: The report does not mention the placement of a carotid stent in the cervical internal carotid artery (ICA). The interventions described are focused on the intracranial segment (M2 segment of the left middle cerebral artery) rather than the cervical segment of the ICA.<br>Conclusion based on the report: Since there is no mention of placing a stent in the cervical ICA, the answer to whether a carotid stent was placed is '0'.<br>The final answer is: 0 |
| Carotid Angioplasty | 0 | To answer the question regarding carotid angioplasty of the cervical Internal Carotid Artery (ICA), we need to review the provided clinical report for any mention of this procedure. The report details an acute stroke intervention where the patient underwent extirpation of an intracranial embolism of the inferior division of the left M2 segment.<br>Checking for cervical ICA angioplasty: The procedure described involves accessing the left internal carotid artery to reach the occluded M2 segment. The report mentions selective catheterization of the left internal carotid artery using a VER catheter and later a VR catheter for repeat angiography. However, it does not explicitly mention angioplasty of the cervical ICA. Angioplasty would typically involve the use of a balloon to dilate a stenosed segment, and this is not described in the provided text.<br>Determining the code: Since there is no mention of performing angioplasty on the cervical ICA, we must conclude that this procedure was not performed. The focus of the intervention was on removing the thrombus from the intracranial segment (M2) using a stentriever and aspiration, without any indication of treating the cervical ICA with angioplasty.<br>The final answer is: 0 |
| Intracranial Stenting | 0 | To determine if an intracranial stent was placed, we need to review the provided clinical report for any mention of stent deployment during the procedure for the stroke patient.<br>Review of Procedure Details: The report mentions the use of a "4 mm x 30 mm TREVO stentriever" during the intervention. The term "stentriever" refers to a device used for mechanical thrombectomy in acute ischemic stroke. It is designed to retrieve clots from occluded cerebral arteries. Although it is called a "stentriever," its primary function is not as a traditional stent that remains in the artery to keep it open but rather as a tool to remove the clot.<br>Understanding Stentriever vs. Traditional Stent: The key distinction here is between a device used for thrombectomy (like the TREVO stentriever) and a traditional intracranial stent, which is left in place to prevent restenosis or to treat a flow-limiting lesion. The report does not mention the deployment of a traditional intracranial stent for the purpose of keeping an artery open.<br>Conclusion Based on Report: Given that the TREVO stentriever is used for clot removal and not left in place as a traditional stent would be, and there is no mention of deploying an intracranial stent for a flow-limiting lesion, the answer to whether an intracranial stent was placed is '0'.<br>The final answer is: 0 |

Table S.4. (continued)

| Variable | Non-Expert Annotation | LLAMA 3.3 Chain-of-Thought Output |
|---|---|---|
| Intracranial Angioplasty | 0 | To determine if intracranial angioplasty was performed, we need to carefully review the provided clinical report for any mention of angioplasty procedures, particularly those targeting intracranial lesions or flow-limiting stenosis.<br>Review of Procedure Details: The report mentions the procedure as "Extirpation of intracranial embolism of inferior division of the left M2" and describes the use of a TREVO stentriever for thrombectomy. This indicates a mechanical thrombectomy was performed to remove a clot, but it does not explicitly mention angioplasty.<br>Device and Technique Description: The report details the use of a Shuttle sheath, a 5Max ace intermediate catheter, a microcatheter, and a TREVO stentriever. These devices are typically used for thrombectomy or stent placement rather than angioplasty. Angioplasty usually involves the use of a balloon to dilate a stenotic segment.<br>Angiographic Findings and Intervention: The findings and interventions described focus on the occlusion of the inferior division of the left M2 segment and the restoration of flow (TICI 2C flow) post-thrombectomy. There is no mention of a flow-limiting stenosis being treated with angioplasty.<br>Conclusion Based on Provided Information: Given the information provided in the clinical report, there is no indication that intracranial angioplasty was performed. The procedures described are related to thrombectomy and the restoration of blood flow, not the dilation of stenotic segments via angioplasty.<br>The final answer is: 0 |
| IV Thrombolytics | 1 | To answer whether IV thrombolysis was administered, we need to review the provided clinical report for any mention of IV tPA (tissue plasminogen activator) or TNK (tenecteplase), which are the medications used for IV thrombolysis in the context of acute stroke.<br>Review of the clinical report: The report mentions that "IV tPA was administered." This directly indicates that the patient received intravenous thrombolysis using tPA (alteplase).<br>2. Determination of IV thrombolysis administration: Since the report explicitly states that IV tPA was given, we can conclude that IV thrombolysis was indeed administered to the patient.<br>3. Coding based on administration: Given that IV tPA was administered, according to the provided instructions, the correct code is '1', which signifies that IV tPA or TNK was given.<br>The final answer is: 1 |
| Site of Occlusion | 3 | To determine the location of the intracranial occlusion, we need to review the provided clinical report for any information regarding the vessel that is occluded.<br>1. Identify which vessel is occluded: The report mentions "Intracranial embolism of inferior division of the left M2" as both the pre-procedure and post-procedure diagnosis. This directly indicates that the occlusion is located in the M2 segment of the left middle cerebral artery (MCA).<br>2. Match to the correct code: Based on the information provided in the question, the codes are as follows:<br>- '1' = ICA (Internal Carotid Artery)<br>- '2' = M1 (1st segment of MCA)<br>- '3' = M2 (2nd segment of MCA)<br>- '4' = Basilar artery<br>- '5' = Other vessel<br>Given that the occlusion is in the M2 segment of the MCA, the correct code is '3'.<br>The final answer is: 3 |

Table S.4. (continued)

| Variable | Non-Expert Annotation | LLAMA 3.3 Chain-of-Thought Output |
|---|---|---|
| TICI Post | 6 | To determine the final TICI score, we need to review the recanalization after endovascular treatment as described in the clinical report provided.<br>1. Review of Recanalization: The report mentions that after the intervention, which included the deployment of a TREVO stentriever and local aspiration with a penumbra, a post-thrombectomy angiogram demonstrated TICI 2C flow. This indicates that the intervention resulted in near-complete perfusion except for small distal branches.<br>2. Decision on TICI Code: The TICI (Thrombolysis In Cerebral Infarction) score is used to grade the flow after thrombectomy or thrombolysis in acute ischemic stroke patients. The scores range as follows:<br>- '0' = No perfusion<br>- '1' = Minimal perfusion<br>- '4' = 2a (partial perfusion <50%)<br>- '5' = 2b (partial perfusion >50%)<br>- '6' = 2c (near-complete perfusion except small distal branches)<br>- '3' = 3 (complete perfusion)<br>Given that the post-thrombectomy angiogram demonstrated TICI 2C flow, which corresponds to near-complete perfusion except for small distal branches, the appropriate TICI code is '6'.<br>The final answer is: 6 |
| NIHSS | 9 | To answer the question regarding the admission NIHSS (National Institutes of Health Stroke Scale) for the stroke patient, we need to follow the chain of thought provided:<br>1. Identify the NIHSS prior to any therapy: The NIHSS is a systematic assessment tool that provides a quantitative measure of stroke-related neurologic deficit. The score is based on the assessment of various neurological functions, with higher scores indicating more severe neurological deficits. The scale ranges from 0 to 42, where 0 indicates no deficit and 42 indicates a severe stroke.<br>2. Must be 0–42: This step reinforces the range within which the NIHSS score can fall, ensuring that any answer provided is within this valid range.<br>3. Given the clinical report provided for the stroke patient, we look for the NIHSS score mentioned:<br>- The report mentions "NIHSS: 9" under the patient's history before any intervention was performed. This indicates that the patient's National Institutes of Health Stroke Scale score was 9 at the time of assessment before undergoing any therapeutic procedures.<br>Therefore, based on the information provided in the clinical report, the admission NIHSS score for this patient is 9.<br>The final answer is: 9 |

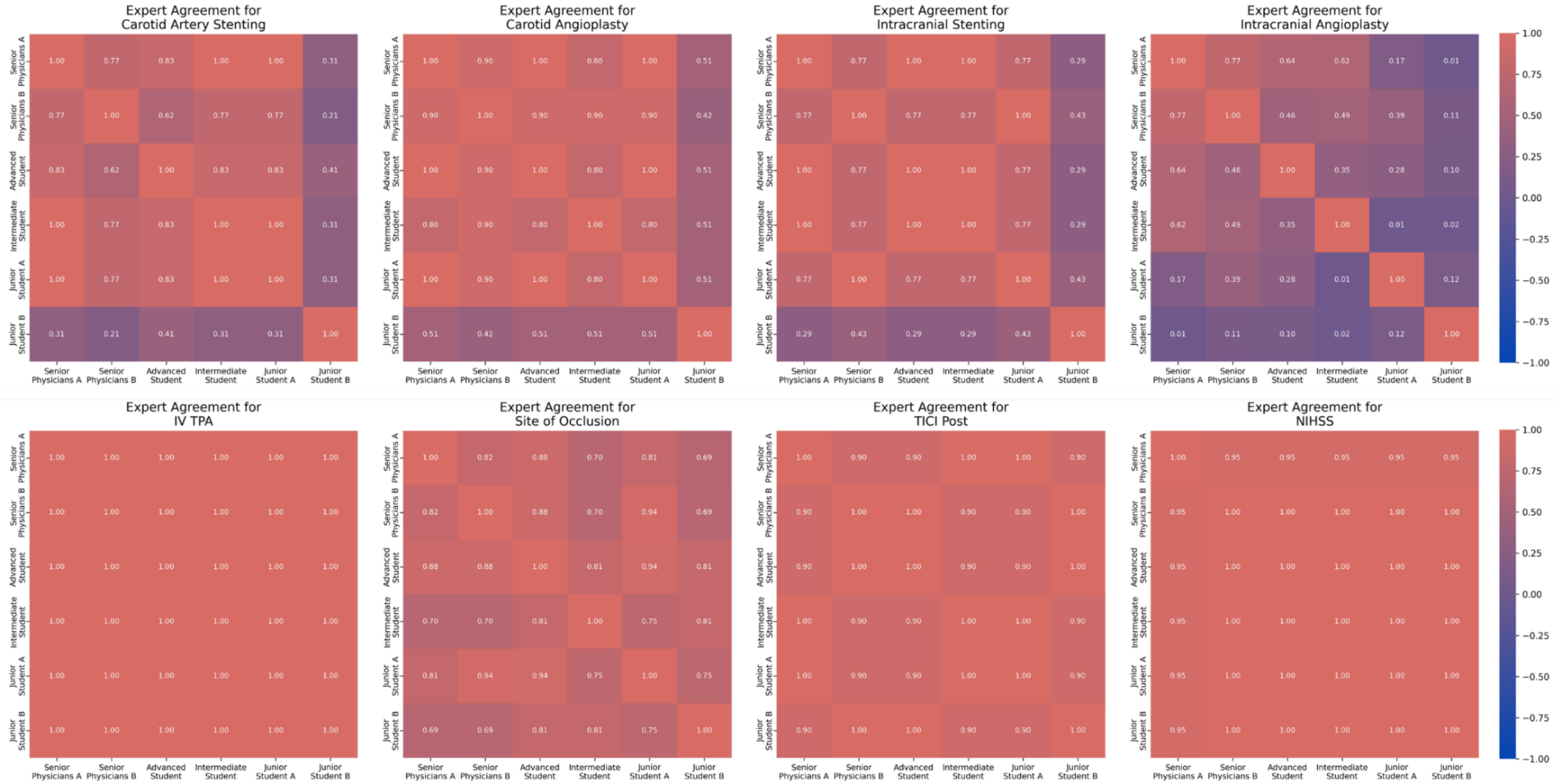

Figure S.1. Heatmaps of pairwise inter-rater agreement for expert adjudicators. Cohen's kappa scores were computed between each pair of adjudicators across eight clinical variables to assess internal consistency in expert labeling. Expert adjudicators showed near-perfect agreement (≈1.00) on most structured variables, reflecting consistent annotation clarity. In contrast, inter-rater reliability dropped for carotid stenting, intracranial stenting, and especially intracranial angioplasty, particularly among junior raters.